\documentclass{article}

\usepackage{graphicx}
\usepackage[numbers]{natbib}

\usepackage{amsmath}
\usepackage{amssymb}
\usepackage{array}
\usepackage{breqn}
\usepackage{mathtools}
\usepackage{hyperref}

\title{Machine Theory of Mind and the Structure of Human Values}
\author{Paul de Font-Reaulx\\
Department of Philosophy\\
University of Michigan, Ann Arbor}

\date{}

\begin{document}

\maketitle
\begin{abstract}
        Value learning is a crucial aspect of safe and ethical AI. This is primarily pursued by methods inferring human values from behaviour. However, humans care about much more than we are able to demonstrate through our actions. Consequently, an AI must predict the rest of our seemingly complex values from a limited sample. I call this the value generalization problem. In this paper, I argue that human values have a generative rational structure and that this allows us to solve the value generalization problem. In particular, we can use Bayesian Theory of Mind models to infer human values not only from behaviour, but also from other values. This has been obscured by the widespread use of simple utility functions to represent human values. I conclude that developing generative value-to-value inference is a crucial component of achieving a scalable machine theory of mind.
\end{abstract}

\section{Introduction}
For artificial agents to be safe and ethical, they need to be able to learn what humans value \cite{christian_alignment_2021, gabriel_artificial_2020, russell_human_2019}. To achieve this, researchers have developed methods such as inverse reinforcement learning for AI to infer a human’s values from their behaviour \cite{arora_survey_2021, ng_algorithms_2000}. These work on the assumption that the person acts to effectively achieve what they want. It is well known that such methods suffer from problems of underdetermination, because many different values can generate the same behaviour \cite{arora_survey_2021, langley_theory_2022, ng_algorithms_2000, quine_word_1960}. One preliminary solution is provided by Bayesian Theory of Mind (ToM) models, which have recently been recognized as a powerful addition to value learning in AI \cite{baker_action_2009, jara-ettinger_theory_2019, langley_theory_2022, mao_review_2023, ruiz-serra_inverse_2023}.

There is, however, a further problem for scalable value learning. Even if an AI is able to reliably and accurately infer human values from human action, there will be many outcomes that we care about but are never able to demonstrate our attitude towards. For example, I value not having a scorpion in my bedroom, but I have never had a chance to evince this value before, say by running away from one. For an AI to learn this value of mine, it will have to predict it from the limited information it has available. The problem is that human values seem complex in a way that makes such generalization difficult, or even intractable, at scale.\cite{russell_human_2019, yudkowsky_fake_2007} If that is correct, it would significantly limit the in-principle scalability of value learning based on inference from behaviour. Call this the \textit{value generalization problem} \cite{arora_survey_2021, langosco_goal_2022}.

In this paper, I argue that the Bayesian ToM of Saxe and others \cite{alanqary_modeling_2021, baker_action_2009, ho_planning_2022, jara-ettinger_naive_2020, zhi-xuan_online_2020, zhi-xuan_solving_2022} can be extended to provide the basis of a solution to the value generalization problem. The crucial observation is that just like actions can be explained as instrumental attempts to realize a value given some beliefs, so other values can be explained as instrumental to realizing more basic values. More generally, our values are connected through a complicated hierarchy of expected value calculations, and this makes them predictable over arbitrary outcomes. This means that we can extend a Bayesian ToM to infer someone’s values not only from their behaviour, but also from their other values. However, this possibility has arguably been neglected because of the widespread use of utility functions that abstract away from the instrumental relations between values. By developing richer representational schemes and models that can represent these instrumental relations, we can overcome the value generalization problem as an impediment to scalable value learning.

In the next section, I provide some background and present the value generalization problem in more detail. In section 3, I develop the claim that human values bear rational relations to each other. In section 4, I consider the implications for inverse reinforcement learning as a value learning technique. I conclude by noting some further problems. Finally, I have added an Appendix with a proposal for how to represent and compute values using causal Bayesian networks.
\section{The Value Generalization Problem}
Here is some background. I take values to be psychological facts. I use the term 'value' because of its use in AI safety, but readers should feel free to exchange it for 'desire' or 'preference'. Formally, we can represent a value as an ordered pair $<o, v>$ consisting of an outcome $o$ and a scalar $v$ representing the agent’s subjective valuation. The valuation is how the agent would judge the outcome if they encountered it, meaning we can value things we have never thought about before. A person's values are the set of $<o, v>$ for any $o$ they care about, which we can represent as a value function $V(o)$.\footnote{I take an outcome to be a set of maximally specific possible worlds. So, for instance, the outcome of the Eiffel tower being in Rome is the set of all possible worlds in which the Eiffel tower is in Rome, each of which has some other variation \cite{jeffrey_logic_1965, stalnaker_inquiry_1984}.} The goal of standard value learning is for a human to communicate or reveal their $V$ to an AI via their behaviour, which in turns forms a representation of the human's values $V'$ \cite{davidson_psychology_1974, ng_algorithms_2000}. For the purpose of this paper, we can assume that the AI has as a goal to maximize $V'$.

The underdetermination problem is that there are many values that are consistent with any given behaviour, meaning that the AI might form an inaccurate representation on the basis of it. For example, suppose that Ben picks a can of anchovies from the store shelf. This behaviour is consistent both with him wanting anchovies, and with him wanting tuna and believing the can to contain tuna. The computational Bayesian ToM developed by Baker, Saxe, and Tenenbaum \cite{baker_action_2009} allows us to make progress on this problem by integrating prior information about the person’s values and beliefs to render a probability distribution across possible values that can best explain the behaviour. For example, if the AI has prior information that Ben hates anchovies, this provides evidence that he chose by mistake.

There is a further issue, however, namely the value generalization problem. Even if Ben is able to communicate his values regarding tuna, van Gogh, and the latest tax raise, and the AI correctly interprets them, there is going to be a large number of outcomes that he never evinces any attitude towards. Aside from many pedestrian things, it will necessarily include outcomes that have never happened before, such as the use of new technology or the occurrence of novel catastrophes. To form a representation of the human’s values about such outcomes, an AI will need to make predictive inferences based on the limited information it has received. More precisely, the set of values that a human is able to communicate $S$ is a subset of $V$. Even if $V'$ agrees with $V$ on $S$, for $V'$ this leaves open the value of any outcome in the domain of $V$ but not in $S$.

If human values were reducible to simple principles—say, that we value things exactly in proportion to how pleasurable and painful they are—then it would be relatively easy to predict how we would value some outcome. What makes the value generalization problem difficult is that human values seem to be irreducible to any such principles, and generally messy. For instance, we value fast cars, burgers, freedom of speech, sex, that Liverpool scores, national holidays, the preservation of the tiger, proving Fermat’s last theorem, and a continual release of HBO series, none of which seem obviously reducible to common denominators of value \cite{yudkowsky_fake_2007}. In AI safety research, this complexity of human values appears widely assumed.

We might propose that we can solve the value generalization problem using statistical techniques. At least in some domains, this does seem to be possible. For example, Spotify is successfully able to predict what music I would value listening to even for bands I have not encountered before using population-level data and correlations to the music I have listened to.\footnote{See also a recent paper by Earp. et al \cite{earp_personalized_nodate} for a fascinating application to predicting the preferences of incapacitated patients.} 

This does not seem like a scalable solution for all domains of human interest. Firstly, most human values are highly contextual. Although we can identify correlations between circumstances and values---e.g. that high income earners are more likely to prefer opera to country---many human situations and potential outcomes will occur in a highly personal context where we should not expect precedents in the population to produce reliable predictions. Also, as Jara-Ettinger \cite{jara-ettinger_theory_2019} observes, statistical approaches to machine ToM require very large amounts of data, which causes problems for outcomes with highly contextual values. Secondly, for outcomes that lack prior instances at all, a statistical approach will necessarily be inadequate, because there will be no precedents to generalize from.

The value generalization problem becomes more severe with the capability of the agent. The more outcomes an AI is able to influence, the more outcomes it needs to have an accurate representation of. If the AI can only affect what music gets selected next in the playlist, then it is a highly limited domains of outcomes that it needs to represent accurately. However, if an AI would develop instrumental capacities exceeding those of humans, then ensuring that it accurately predicts our values over even far-fetched catastrophic scenarios becomes more acute \cite{bostrom_superintelligence_2014}. Unfortunately, those scenarios are also those we should expect statistical techniques to be especially unreliable for, given their lack of precedent. Therefore, finding another solution to the value generalization problem is crucial to long-term AI safety.

\section{The Rational Structure of Human Values}
Despite its apparent prevalence in value learning research, the claim that human values are irreducibly complex is false. Our values have a rational generative structure that makes them predictable. The clearest illustration of this is that we humans can predict each other’s values even over outcome that the other person might never have considered, and for people we barely know. For example, I don’t know Emmanuel Macron’s cousin, but I am highly confident that he values not having a scorpion in his bedroom tonight. I could not have learnt this from his behaviour towards scorpions, since I have never interacted with him. Rather, I confidently believe he would find it detrimental to other things I have good reason to think he values, such as health and bodily integrity, and infer that he would attribute low value to it.

The instrumental rationality assumption that we use to infer a value from an action is true not only between actions and values, but also between values themselves. Philosophers express this by saying that some values are instrumental to other outcomes we value intrinsically \cite{schroeder_three_2004, tiberius_well-being_2018}. For example, I might value being vaccinated against the flu. But I value it instrumentally because achieving that outcome improves my chances of satisfying my more basic value of not having the flu.\footnote{As Hume puts it in the second enquiry: "Ask a man, why he uses exercise; he will answer, because he desires to keep his health. If you then enquire, why he desires health, he will readily reply, because sickness is painful. If you push your enquiries farther, and desire a reason, why he hates pain, it is impossible he can ever give any. This is an ultimate end, and is never referred to any other object." \cite{hume_enquiry_1983}}

The claim that our values have this instrumental structure is not just philosophical speculation, however. It also has overwhelming empirical support from cognitive neuroscience \cite{dayan_reinforcement_2008, glimcher_neuroeconomics_2013, krajbich_visual_2010, montague_imaging_2006, rangel_framework_2008, serra_decision-making_2021}. There is now a plethora of studies showing that many cognitive functions and general motivation is best explained in terms of expected value computations performed in our minds over a basic currency of subjective value. Such expected value calculations are just a more general and formal representation of the instrumental cognitive structure articulated above.

This structure means that we can generatively infer new values from other values of an agent that we already know. For example, if I know that Miriam values not having the flu and believes that being vaccinated will help with that, then I can infer that she will value being vaccinated, and that probably get a flu shot. Conversely, if I see Miriam getting a flu shot I can infer that she values being vaccinated. And once I am confident that she has this value, I can infer that she values not having the flu.

See Figure 1 for an illustration of this example. The direction of reasoning is the causal reasoning of Miriam, and the direction of explanation the series of answers we get when we ask why she does or values something. The arrows between the shapes represent causal relations in Miriam's world model. The rectangle is an action and the circles outcomes, though either can be represented as being in the domain of $V$. Notice that this is distinct from traditional decision networks in AI \cite{norvig_artificial_2021}, but can be represented using Jeffrey-Bolker decision theory \cite{jeffrey_logic_1965}.

\begin{figure}
  \centering
  \includegraphics[width=10cm]{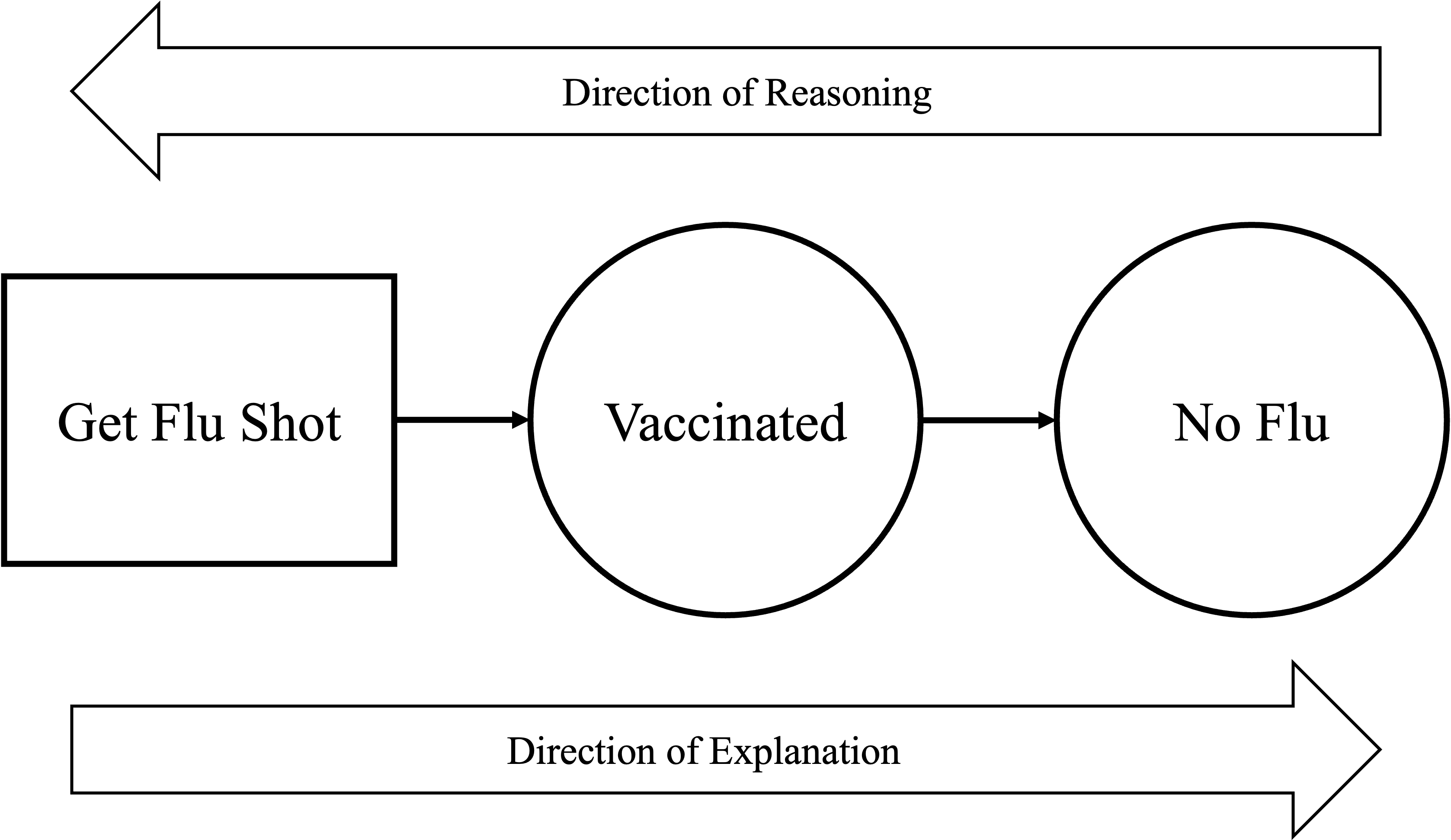}
  \caption{An illustration of Miriam's action and values, and their instrumental relations.}
\end{figure}

We can express the relation between Miriam's values more precisely. In particular, the value of Miriam's $V(Vaccinated)$ is a function of how much she expects that state to make a difference to her chances of getting the flu on her world model. In other words, assuming that there is no intrinsic value to being vaccinated, it inherits its value from its expected impact on the more fundamentally valued outcome of not having the flu. In the Appendix, I provide a general formula to compute values like this in general using causal Bayesian networks.\footnote{In this case, $V(Vaccinated)$ is given by the difference in flu risk---weighed by the negative value of having relative to not having the flu---between being vaccinated and not being vaccinated, i.e. between $V(No\ Flu)P(No\ Flu|Vaccinated) + V(Flu)P(Flu|Vaccinated)$ and $V(No\ Flu)P(No\ Flu|\neg Vaccinated) + V(Flu)P(Flu|\neg Vaccinated)$.}

These are trivial applications of a Bayesian ToM. But they importantly illustrate that the same kind of interpretative inferences can be made between values as between action and value. Rather than requiring behaviour for any outcome to know how a person values that outcome, this structure allows us to infer other values from those that we accurately elicit. If we imagine a person’s values as a jigsaw puzzle to be solved, observing the rational structure of our values allows us to fill in the picture by seeing what pieces fit with each other.

Importantly, the instrumental structure of our values extends even to unobserved outcomes, such as unprecedented catastrophes. This is why we humans can reliably predict that other humans would value that such catastrophes do not occur, even though no human has signalled a value about such an outcome before: it would be deeply detrimental to outcomes we care more fundamentally about. This is good news for long-term AI safety concerns, because it means that a normal human attitude towards large scale catastrophe should be easy to infer from our other values, if the artificial agent is equipped with a generative theory of mind that allows for predictive inferences between instrumental values. In other words, the rational structure of our values allows for far more scalable value learning than seems commonly accepted today.\footnote{The point made here is similar to that of Ho et al. \cite{ho_planning_2022}, though they focus on planning for action rather than relations between values. Importantly, the point can also be expressed in terms of world models in model-based reinforcement learning.}

\section{Inverse Reinforcement Learning and the Representation of Human Values}

The observation that I can value outcome $A$ because I value outcome $B$, and believe that $A$ would be instrumental to $B$, is a simple one that surprisingly has not been properly appreciated in the context of value learning. I believe this is a consequence of how we represent human values in discussions of inverse reinforcement learning (IRL), which is the primary paradigm of value learning for AI.

In IRL we represent the human agent as a reinforcement learning (RL) agent with a so-called reward function representing their subjective valuation of outcomes. Using RL-algorithms, an IRL-agent tries to infer what the human’s reward function is from their behaviour, and then maximize that. (Note that this is just an instance of the rational interpretation method of value learning we have been discussing above.) It is standard to treat human reward as equivalent to a human utility function, merely expressed in terms of an alternative formalism \cite{arora_survey_2021, ruiz-serra_inverse_2023}. I believe this approach to value learning gives rise to two significant problems which preclude and distort information about the empirical structure of human values, in a way that handicaps the potential for scalable value learning.

Firstly, a utility function in the microeconomic sense is merely a numerical way of representing a preference ranking \cite{mas-colell_microeconomic_1995, savage_foundations_1954}. In other words, if the utility of $A$ is more than that of $B$ for an agent, this just says that the agent values $A$ more than $B$ (and if the utility function is cardinal, how much more). It does not—and cannot—represent facts such as whether the agent values $A$ because of its expected contribution to $B$. For example, using only a utility function we can say that Miriam values the outcomes of being vaccinated and the outcome of being flu-free, but we cannot say that she values being vaccinated \textit{because} she values being flu-free. This means that representing human values in terms of utility functions loses a lot of important information by projecting the hierarchical structure of our values onto a single-dimensional measure of comparative subjective value.

Secondly, treating this utility function as a human reward function leads to a conflation of reward and value in the psychology of humans. On a modern orthodox account of human motivation, we—like any learning agent—have a capacity to evaluate experiences as positive or negative, which is a function mapping states to a scalar representing basic subjective value \cite{haas_reinforcement_2022, sutton_reinforcement_2018, tomasello_evolution_2022}. This could for example be whether they are pleasurable or painful, although that particular suggestion has turned out to be empirically implausible \cite{schroeder_three_2004}. In an RL-framework, this would be the reward function. We then compute the total value of an outcome using predictive algorithms trying to estimate its long-term effects, given some plan of action. The outcome of these computations are the valuations of outcomes we have represented in terms of a value function $V$. We can equate $V$ with a standard cardinal utility function. However, notably $V$ is not reward in human psychology, but the total subjective value of an outcome. In other words, strictly speaking IRL ends up eliciting human preferences; not the human reward function that was a part of originally generating those preferences.

Consider how IRL might go wrong in Miriam's case. Observing that Miriam chooses to receive a vaccine, an IRL agent might naturally infer that Miriam finds Vaccinated a rewarding state, i.e. that her reward function attributes positive value to it. This leads to two problems corresponding to the issues above. First, it misrepresents her psychology. Miriam does not choose to get the shot because being vaccinated is rewarding, but because it is instrumentally valuable in expectation. Second, because there is no representation of the expected instrumental relation between being vaccinated and being flu-free, there is no way to make the further inference that Miriam values being flu-free.

Future effective value learning will require a framework that allows us to express not only that we value an outcome, but why we value that outcome. An example of such a framework is a Bayesian network representing the relations between outcomes on an agent’s predictive models, such as illustrated in the example of Miriam above. Furthermore, the RL-framework does also provide the basic tools to represent these relations in terms of world models in model-based RL \cite{sutton_reinforcement_2018}. If models based on such frameworks could be integrated into IRL algorithms, that could constitute a solution to the value generalization problem and the basis for a scalable value learning in AIs.\footnote{Note that there are several examples of Bayesian approaches to IRL \cite{choi_map_2011, lopes_active_2009, ramachandran_bayesian_2007}, but none to my knowledge that integrate hierarchical value structures. }

\section{Conclusion}

In this paper, I have argued that human values are rationally structured in a way that is neglected in value learning research, and which allows for generative inferences of human values from other values using, for example, the Bayesian theory of mind methods developed by Saxe and others. This constitutes an important component of developing scalable value learning for safe and ethical AI.

Significant questions remain, however. Suppose for instance that an AI infers that a human values an outcome on the basis of false beliefs. Should the AI prioritize the outcome that the human in fact wants, or what they would have wanted had they had more accurate beliefs? For example, suppose that a person values smoking because they falsely believe it has negligible effects on their health, and would not otherwise smoke. Should an AI facilitate their smoking? \cite{railton_moral_1986, williams_internal_1979} Developing ethical and safe AI that is able to handle such questions will require a close cooperation between engineering and philosophy going forward.

\section*{Appendix: A Proposal for Computing Value}
In the example above we saw a simple explanation of how to compute the value of the outcome $Vaccinated$ from its expected impact on $No\ Flu$. In this section I briefly present a general formula for computing the value of an outcome.

\begin{figure}
  \centering
  \includegraphics[width=4cm]{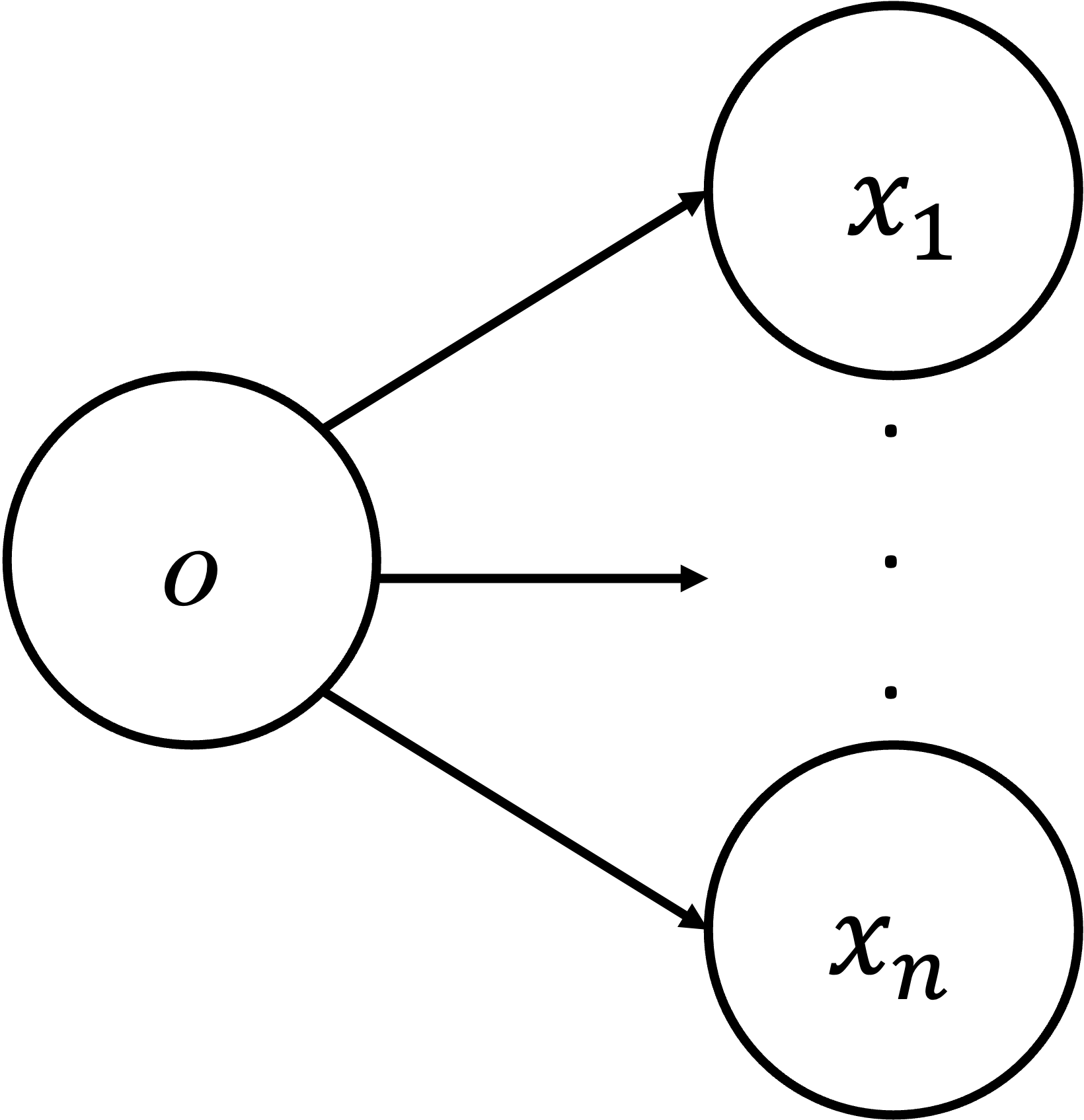}
  \caption{A world model with expected causal impact between $o$ and $x_1,...x_n$.}
\end{figure}

Consider an agent with a world model such that an outcome $o$ causally bears on outcomes $x_1,...x_n$. See this illustrated in Figure 2. Let $r(o)$ represent the intrinsic value of $o$, and $P(\cdot)$ the agent's subjective probability function. $V(o)$ is the sum of the intrinsic value of $o$ and the difference it makes to the probabilities of other outcomes $x_1,...x_n$. Using an $Impact$ operator to represent this difference-making, the formula for $V(o)$ is:

\[V(o)=r(o)+\sum_{i=1}^{n} Impact(o, x_i)\]

The $Impact$ operator represents the difference in expected value from possible outcomes if $o$ were to occur relative to if it were to not occur. Intuitively, the bigger $o$ is as a positive difference-maker according to the world-model of the agent, the more instrumentally valuable it is. For example, if someone were very sensitive to the flu, then the value of being vaccinated would be great. But if they were immune regardless of being vaccinated, then the value would be negligible. Here is a simple version of $Impact$:

\begin{align*}
        Impact(o, x)=\overbrace{[V(x)P(x|o)+V(\neg x)P(\neg x|o)]}^{Expected\ value\ of\ o\ w.r.t.\ x}-\\
        \underbrace{[V(x)P(x|\neg o)+V(\neg x)P(\neg x|\neg o)]}_{Expected\ value\ of\ \neg o\ w.r.t.\ x}
\end{align*}

This version is simplified in two ways. First, we should use Pearl's $do$ operator to replace $o$ with $do(o)$ \cite{pearl_causality_2009}. The $do$ operator is usually used to represent action, but here we use it to represent hypothetical intervention in an agent's world model. In particular, this rules out updates to the probability of any parent nodes of $o$ (not included in Figure 2). Second, to insure causation we should replace the conditional probabilities with probabilities of subjunctive conditionals \cite{stalnaker_theory_1968}. For example, we should replace $V(x)P(x|do(o))$ with $V(x)P(do(o)>x)$), where $P(A>B)$ means the probability that if $A$ were to occur then $B$ would occur. This is distinct from the probability of $B$ given $A$, which does not imply a causal connection.

Finally, it is worth noting that the proposed formula is structurally similar to the Bellman equation for calculating the value of a state using the value iteration algorithm in RL \cite{sutton_reinforcement_2018}. However, setting aside the difference in framing, the primary difference is that value iteration is used to compute the total value of the current state, while $Impact$ operator subtracts the value that is not attributable to $o$. For example, using value iteration, the value of $Vaccinated$ would be the same whether or not you were previously immune to the flu, because what matters is just the probability of getting the flu. By contrast, in the formula above, the value of $Vaccinated$ is higher if you are not already immune, because only then does it make a difference.
\bibliography{Bib_2}

\end{document}